\documentclass[runningheads]{llncs}
\usepackage{graphicx}
\usepackage{epsfig,psfig}
\usepackage{amssymb}
\begin{document}

\title{Enhancing Indic Handwritten Text Recognition using Global Semantic Information}


\author{Ajoy Mondal\orcidID{0000-0002-4808-8860} \and
C. V. Jawahar\orcidID{0000-0001-6767-7057}}

\authorrunning{Ajoy Mondal and C. V. Jawahar}

\institute{International Institute of Information Technology, Hyderabad, India \\
\email{ajoy.mondal@iiit.ac.in and jawahar@iiit.ac.in}}

\maketitle  

\begin{abstract}

Handwritten Text Recognition ({\sc htr}) is more interesting and challenging than printed text due to uneven variations in the handwriting style of the writers, content, and time. {\sc htr} becomes more challenging for the Indic languages because of (i) multiple characters combined to form conjuncts which increase the number of characters of respective languages, and (ii) near to 100 unique basic Unicode characters in each Indic script. Recently, many recognition methods based on the encoder-decoder framework have been proposed to handle such problems. They still face many challenges, such as image blur and incomplete characters due to varying writing styles and ink density. We argue that most encoder-decoder methods are based on local visual features without explicit global semantic information.

In this work, we enhance the performance of Indic handwritten text recognizers using global semantic information. We use a semantic module in an encoder-decoder framework for extracting global semantic information to recognize the Indic handwritten texts. The semantic information is used in both the encoder for supervision and the decoder for initialization. The semantic information is predicted from the word embedding of a pre-trained language model. Extensive experiments demonstrate that the proposed framework achieves state-of-the-art results on handwritten texts of ten Indic languages. 
\end{abstract}

\section{Introduction}

Optical Character Recognition ({\sc ocr}) is the electronic or mechanical conversion of printed or handwritten document images into a machine-readable form. {\sc ocr} is an essential component in the workflow of document image analytics. Typically, an {\sc ocr} system includes two main modules (i) a text detection module and (ii) a text recognition module. A text detection module aims to localize all text blocks within the image, either at the word or line levels. The text recognition module aims to understand the text image content and transcribe the visual signals into natural language tokens. The problem of handwritten text recognition is more interesting and challenging than printed text due to the presence of uneven variations in handwriting style to the writers, content, and time. The handwriting of a person is always unique, and this uniqueness property creates motivation and interest among the researchers to work in this exigent and challenging field. 

Among various languages around the world, many of them have disappeared as their usage is limited due to their presence in rural or geographically inaccessible parts of the globe. At this point, it is highly recommended to use technologies like {\sc ocr} and natural language processing to stop the extermination of languages in the world. There are almost 7000 languages in the world\footnote{\url{https://www.ethnologue.com/guides/how-many-languages}}. However, handwritten {\sc ocr} systems/tools are available only for a few. 
{\sc ocr} systems are mostly available for languages that are of huge importance and strong economic value, like English~\cite{graves2008offline,pham2014dropout,li2021trocr}, Chinese~\cite{xie2016fully,wu2017handwritten,peng2022recognition}, Arabic~\cite{maalej2020improving,jemni2022domain}, and Japanese~\cite{ly2018training,nguyen2020semantic}. Most of the languages derived from Indic script appear to be at the risk of vanishing due to the absence of efforts. So, there is an immense need for character recognition-related research for Indic scripts/languages.

Officially, there are 22 languages in India, many of which are used only for communication purposes. Among these languages, Hindi, Bengali, and Telugu are the top three languages in terms of the percentage of native speakers~\cite{krishnan2019hwnet}. In many Indic scripts, two or more
characters are often combined to form conjuncts which considerably increase the number of characters/symbols to be tackled by {\sc ocr} systems~\cite{citekey}. These inherent features of Indic scripts make Handwritten Recognizer ({\sc hwr}) more challenging as compared to Latin scripts. Compared to the 52 unique (upper case and lower case) characters in English, most Indic scripts have over 100 unique basic Unicode characters~\cite{pal2004indian}.

The problem of Indic handwritten text recognition, more generally text recognition, is formulated as a seq-2-seq prediction task where both the input and output are treated as a sequence of vectors. It aims to maximize the probability of predicting the output label sequence given the input feature sequence~\cite{adak2016offline,dutta2017towards,dutta2018towards}. The encoder-decoder (e.g., {\sc cnn}-{\sc rnn}) framework is very popular to handwritten text recognition tasks~\cite{dutta2017towards,dutta2018towards,gongidi2021iiit}. The encoder extracts rich features and generates a context vector containing global information of the input text image. While the decoder converts the context vector to the target text.     

\begin{figure}
\centerline{
\includegraphics[width=1.0\textwidth]{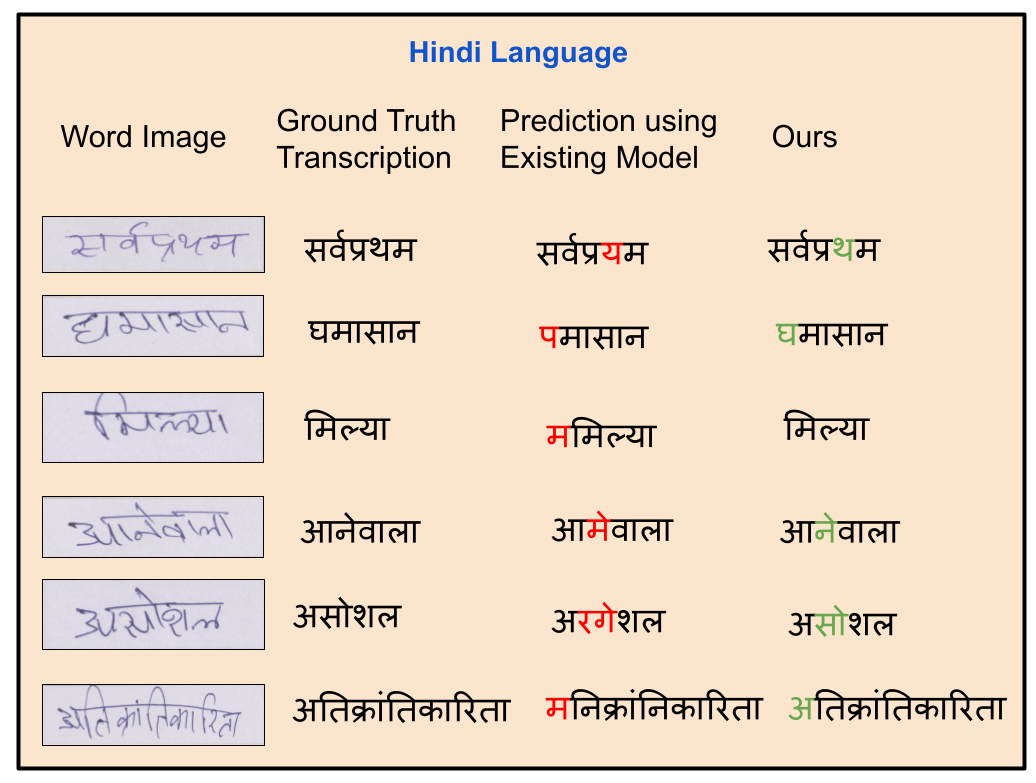}}
\caption{Shows a comparison of our method with the existing encoder-decoder framework~\cite{gongidi2021iiit}. The first column shows examples of some 
challenging Hindi word images. The second column is ground truth transcription. The third column is the results of method~\cite{gongidi2021iiit}. The fourth column gives the predictions of our approach. The Red colored characters are wrongly recognized by method~\cite{gongidi2021iiit}. The Green colored characters are correctly recognized by the proposed method.\label{fig:image_sample}}
\end{figure}

Despite great effectiveness, the encoder-decoder framework has limited ability to generate context information that represents the whole input image~\cite{das_2022}. Inspired by human visual attention, the researchers introduce the attention mechanism into the encoder-decoder framework, referred to as the attention-based encoder-decoder framework. Several works~\cite{shi2018aster,wang2020decoupled,das_2022} exist in this direction. The attention mechanism helps the decoder to select the appropriate context at each decoding step resulting in accurate text recognition. It can resolve long-range dependency problems and the alignment between the encoder and decoder module. Kass and Vats~\cite{das_2022} show that the attention mechanism improves accuracy more than a simple encode-decode framework for recognizing handwritten English text. This type of framework works accurately for most of the scenarios except for low-quality images. In the case of handwritten text, due to varying handwriting styles and the density of the ink, text may be distorted, blurred, and have incomplete characters. Global semantic information is an alternative feature to handle these problems.

In this work, we enhance the performance of Indic handwritten text recognizers using global semantic information. Inspired by the work~\cite{qiao2020seed}, we propose an Indic handwritten text recognizer based on an attention-based encoder-decoder framework with an additional semantic information module to predict global information. The semantic information is used to initialize the decoder and it has two main advantages (i) it can be supervised by a word embedding in the natural language processing field and (ii) it can reduce the gap between the encoder focusing on the visual feature and the decoder focusing on the language information. The training goal is to reduce the difference between the predicted semantic information and word embedding from a pre-trained language model. This way, the semantic module predicts richer semantic information, guides the decoder during the decoding process, and improves decoder performance. Some examples are shown in Fig.~\ref{fig:image_sample}. However, our framework can correct it with the global semantic information. In other words, semantic information works as an “intuition”, like a glimpse before people read a word carefully.

The main contributions of this work are as follows

\begin{itemize}
\item We propose an attention-based encoder-decoder framework with a semantic module to recognize Indic handwritten text. The semantic module predicts global semantic information, which guides the decoder to recognize text accurately. FastText, the pre-trained language model, supervises the predicted semantic information. 

\item Extensive experiments on public benchmark Indic handwritten datasets demonstrate that the proposed framework obtains state-of-the-art performance.     
\end{itemize}

\section{Related Work}

There are three popular ways of building handwritten word recognizers for Indic scripts in the literature. The first one is to use segmentation-free, but lexicon-dependent methods train
on representing the whole word~\cite{shaw2008offline,shaw2014combination,kaur2021recognition}. In~\cite{shaw2008offline}, Shaw {\em et al.} represent word images using a histogram of chain-code directions in the image strips scanning from left to right by a sliding window as the feature vector. A continuous density Hidden Markov Model ({\sc hmm}) is proposed to recognize handwritten Devanagari words. Shaw {\em et al.}~\cite{shaw2014combination} discuss a novel combination of two different feature vectors for holistic recognition of offline handwritten word images in the same direction. 

Another approach is based on segmentation of the characters within the word image and recognition of isolated characters using an isolated symbol classifier such as Support Vector Machine ({\sc svm})~\cite{arora2010performance}, Artificial Neural Network ({\sc ann})~\cite{labani2018novel,alonso2014combining}. In~\cite{roy2016hmm}, Roy {\em et al.} segment Bengali and Devanagari word images into the upper, middle, and lower zones, using morphology and shape matching. The symbols present in the upper and lower zone are recognized using an {\sc svm}, while a Hidden Markov Model ({\sc hmm}) was used to recognize the characters in the middle zone. Finally, the results from all three zones are combined. This category of approaches suffers from the drawback that we have to use a script-dependent character segmentation algorithm. 

The third category of approaches treats word recognition as a seq-2-seq prediction problem where both the input and output are treated as a sequence of vectors. The aim is to maximize the probability of predicting the output label sequence given the input feature sequence~\cite{adak2016offline,dutta2017towards,dutta2018towards}. Garain {\em et al.}~\cite{garain2015unconstrained} proposed a recognizer using Bidirectional Long short-term memory ({\sc blstm}) with Connectionist Temporal Classification ({\sc ctc}) layer to recognize unconstrained Bengali offline handwriting words. Adak {\em et al.}~\cite{adak2016offline} used Convolutional Neural Network ({\sc cnn}) integrated with an {\sc lstm} along with a {\sc ctc} layer to recognize Bengali handwritten words. In the same direction, Dutta {\em et al.} proposed {\sc CNN-RNN} hybrid end-to-end model to recognize Devanagari, Bengali~\cite{dutta2017towards}, and Telugu~\cite{dutta2018towards} handwritten words. In this work~\cite{gongidi2021iiit}, the authors use Spatial Transformer Network along with hybrid {\sc cnn-rnn} with {\sc ctc} layer to recognize word images in eight different Indic scripts such as Bengali, Gurumukhi, Gujarati, Odia, Kannada, Malayalam, Tamil, and Urdu. The authors use various data augmentation functions to improve recognition accuracy. This category of methods does not require character-level segmentation and is not bounded for recognizing a limited set of words. 

\section{Proposed Method}

\begin{figure}
\centerline{
\includegraphics[width=1.0\textwidth]{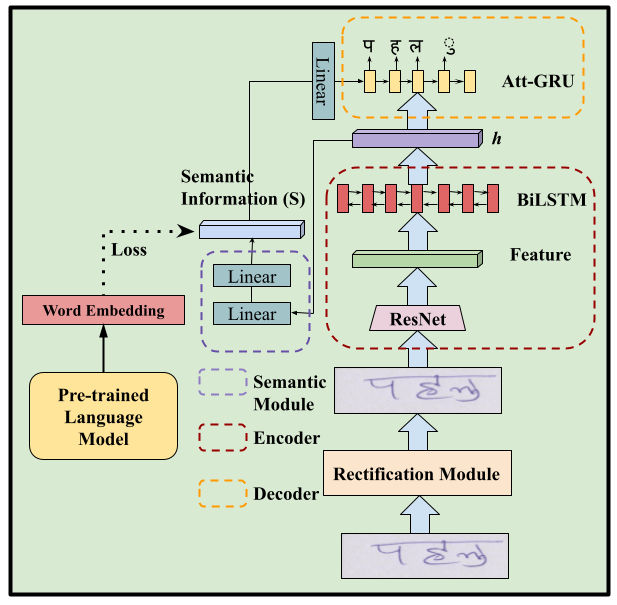}}
\caption{Presents detail of the proposed Indic handwritten text recognizer. It consists of four main components: rectification module, encoder, semantic module, and decoder.}\label{fig:architecture}
\end{figure}

The proposed method for Indic handwritten text recognition tasks is discussed in detail. Fig.~\ref{fig:architecture} shows the proposed Indic handwritten text recognizer, which consists of four components: (i) the rectification module to rectify the irregular word image, (ii) the encoder to extract rich visual features, (iii) the semantic module to predict semantic information from the visual feature, and (iv) the decoder transcribes the final recognized text.

Most of the text recognizers are built on the encoder-decoder architecture with attention. The decoder focuses on specific regions of visual features extracted by the encoder and recognizes the corresponding characters step by step. This type of framework works accurately for most of the scenarios except for low-quality images. In the case of handwritten text, due to varying handwriting styles and the density of the ink, text may be distorted and blurred. Global semantic information is an alternative feature to handle these problems. As shown in Fig.~\ref{fig:architecture}, the semantic module extracts semantic information, which helps the decoder to predict accurate characters. The use of word embedding from a pre-trained language model acts as a supervisor to extract semantic information and help the decoder to improve its performance. In the following subsections, we discuss each module in detail. 

\subsection{Rectification Module}

Generally, the diverse handwriting style present in handwritten data makes handwritten text recognizers more challenging. The rectification module learns to apply input-specific geometric transformations to rectify the image. By rectifying the input image, this module simplifies the recognition tasks. Normally Spatial Transformer Network ({\sc stn})~\cite{jaderberg2015spatial} and its variants are used to rectify the input text image. In this work, the rectification module is based on {\sc stn}. The {\sc stn} models spatial transformation as the learnable parameters for the given input text image. The network first predicts a set of control points via its localization network. Then Thin-plate Spline ({\sc tps})~\cite{bookstein1989principal} transformation is calculated from the control points and passed to the grid generator and the sampler to generate the rectified image. Since the control points are predicted from the input text image, the rectification network takes no extra inputs other than the input image. 

\subsection{Encoder Module} 

The rectified image is forwarded to the encoder module consisting of a 45-layer ResNet based {\sc cnn} similar to~\cite{shi2018aster} followed by a 2-layer Bidirectional {\sc lstm} ({\sc b}i{\sc lstm})~\cite{graves2008novel}. The output of the encoder module is a feature sequence $h=(h_{1}, ..., h_{L})$ with the shape of $L\times C$, where $L$ is the width of the last feature map in {\sc cnn} and $C$ is the depth.

\subsection{Semantic Module}

The semantic module uses feature sequence $h$ to predict semantic information for additional input to the decoder for accurately predicting characters. For this purpose, we flatten the feature sequence $h$ into a one-dimensional feature vector $X$ with the dimension of $K$, where $K=L\times C$. The semantic module predicts semantic information $S$ through two linear functions as follows 
\begin{equation}
S = W_{2} \sigma(W_{1}X + b_{1}) + b_{2}.     
\end{equation}
Where $W_{1}$, $W_{2}$, $b_{1}$, and $b_{2}$ are trainable parameters of the linear function and $\sigma$ (i.e., ReLU) is an activation function. Word embedding from pre-trained language model (e.g., FastText model) acts as a supervisor to predict semantic information.    

\paragraph{\textbf{FastText Model:}}

In this work, we use word embedding based on skip-gram from FastText~\cite{bojanowski2017enriching}, a pre-trained language model. In a text corpus, suppose $T={w_{i}-1, ..., w_{i}+1}$ be a sentence and $l$ indicates its length. A word $w_{i}$ is represented by a single embedding vector $v_{i}$ in skip-gram. The embedded representation $v_{i}$ of a word $w_{i}$ inputs to a simple feed-forward neural network and predicts the context representation as $C_{i}=\{w_{i}-l, . . ., w_{i}-1, w_{i}+1, . . ., w_{i}+l \}$. The input embedding vector $v_{i}$ is simultaneously optimized through the feed-forward network training. Finally, the optimized embedding vector of a word is very close to the words with similar semantics.         
FastText embeds subwords and uses them to generate the final embedding of the word $w_{i}$. Two hyper-parameters $l_{min}$ and $l_{max}$ denote minimum and maximum lengths of subwords. For example, in the word "which" with $l_{min}=2$ and $l_{max}=4$, the possible subwords are {wh, wi, wc, hi, hc, hh, ic, ih, ch, whi, whc, whh, hic, hih, ich, whic, whih, hich}. Embedding vectors of all subwords are combined to represent the corresponding word.  

\subsection{Decoder Module}

The decoder consists of a single layer attentional {\sc gru}~\cite{cho2014learning} with 512 hidden units and 512 attention units. The decoder adopts the Bahdanau-Attention mechanism~\cite{bahdanau2014neural}. The decoder is single-directional. The semantic information $S$ is used to initialize the state of {\sc gru}. The decoder uses both local visual information $h$ and global semantic information $S$ to generate more accurate results.  

\subsection{Loss Functions and Training Strategy}

The proposed model is trained in an end-to-end manner. We add supervision in both the semantic module and the decoder module. The loss function is defined as 
\begin{equation}
L = L_{r} +\lambda L_{e},
\end{equation}
where $L_{r}$ is the standard cross-entropy loss of the predicted probabilities with respective ground truths and $L_{e}$ is the cosine embedding loss of the predicted semantic information $S$ to the word embedding of the
transcription label from the pre-trained FastText model. $\lambda$ is a balancing parameter, and in this work, we set it as $1$. Cosine embedding loss $L_{e}$ is defined as $L_{e} = 1-cos(S,\ E)$, where $S$ is predicted semantic information through a semantic module, and $E$ is the word embedding from the pre-trained FastText model.  

\subsection{Implementation Details}

The proposed Indic handwritten text recognizer is implemented in PyTorch (base code is available in~\footnote{\url{https://github.com/Pay20Y/SEED}}). We use pre-trained FastText model~\footnote{\url{https://fasttext.cc/docs/en/crawl-vectors.html}} trained on common Crawl~\footnote{\url{https://commoncrawl.org/}} and Wikipedia~\footnote{\url{https://www.wikipedia.org}}. We use 45-layer ResNet architecture similar to~\cite{shi2018aster} and 2-layer Bidirectional {\sc LSTM} with $256$ hidden units. We resize the input text image into $64\times256$ without keeping the ratio. We train the model for $50$ epochs and use Adadelta~\cite{graves2008novel} to minimize loss functions. We set the batch size to $64$ and the learning rate to $1.0$. 

For inference, we resize the input images to the same size, similar to the training process. We use beam search for {\sc gru} decoding and set beam width to $5$ for all experiments. 

\paragraph{\textbf{Data Augmentation:}} The works~\cite{dutta2018improving,dutta2018towards,wigington2017data,gongidi2021iiit} highlight that the data augmentation strategies improve the handwritten text recognizer performance on Latin and Indic scripts. It enables the network to learn invariant features for a given task and prevents over-fitting. Similar to the work~\cite{gongidi2021iiit}, we apply affine and elastic
transformations on input text images to imitate the natural distortions and variations presented in handwriting text. We also use brightness and contrast augmentation on input text images to learn invariant features for text and background.

\section{Experiments}

\subsection{Datasets}

We used publicly available benchmark Indic handwriting dataset: {\sc iiit-indic-hw-words}~\cite{dutta2018offline,dutta2018towards,gongidi2021iiit} for this experiment. Table~\ref{table_dataset} shows details of the used datasets containing word images of ten different languages. 

\begin{table*}[ht!]
\addtolength{\tabcolsep}{7.0pt}
\begin{center}
\begin{tabular}{|l|r|r|r|} \hline
\textbf{Script} &\textbf{\#Writer} &\textbf{\#Word Instance} &\textbf{\#Lexicon} \\ \hline
Bengali   &24   &113K &11,295 \\ 
Gujarati  &17   &116K &10,963 \\
Gurumukhi &22   &112K &11,093 \\
Devanagari  &12   &95K &11,030 \\
Kannada   &11   &103K &11,766 \\
Odia      &10   &101K &13,314 \\
Malayalam &27   &116K &13,401 \\
Tamil     &16   &103K &13,292 \\
Telugu    &11   &120K &12,945 \\ 
Urdu      &8    &100K &11,936 \\ \hline
\end{tabular}
\end{center}
\caption{Shows the statistics of used datasets for this experiments. 
\textbf{\#:} indicates number. \label{table_dataset}} 
\end{table*}

\subsection{Evaluation Metric}

Two popular evaluation metrics such as Character Recognition Rate ({\sc crr}) (alternatively Character Error Rate, {\sc cer}) and Word Recognition Rate ({\sc wrr}) (alternatively Word Error Rate, {\sc wer}) are used to evaluate the performance of recognizers. Error Rate ({\sc er}) is defined as 

\begin{equation}
ER = \frac{S + D + I}{N}, \label{equation1}
\end{equation}
where $S$ indicates number of substitutions, $D$ indicates the number of deletions, $I$ indicates the number of insertions and $N$ the number of instances in reference text. In case of {\sc cer}, Eq.~(\ref{equation1}) operates on character levels, and in case of {\sc wer}, Eq.~(\ref{equation1}) operates on word levels. Recognition Rate ({\sc rr}) is defined as 
\begin{equation}
RR = 1-ER. \label{equation2}
\end{equation}
In the case of {\sc crr}, Eq.~(\ref{equation2}) operates on character levels and in the case of {\sc wrr}, Eq.~(\ref{equation2}) operates on word levels.

\subsection{Results}

\paragraph{\textbf{Ablation Study}}
We perform three different experiments for our ablation study such as (i) only encoder-coder framework, (ii) encoder-decoder with attention module, and finally (iii) encoder-decoder with attention and semantic modules. There are two steps to the semantic module in the proposed method such as (i) word embedding supervision and (ii) initialization of decoder with predicted semantic information. We evaluate these two steps separately by using {\sc iiit-indic-hw-words}~\cite{gongidi2021iiit} as a training dataset. Table~\ref{table_ablation_study} shows the results of the handwritten word dataset of the Bengali script. 

\begin{table}[ht!]
\addtolength{\tabcolsep}{7.0pt}
\begin{center}
\begin{tabular}{|l|l| r r r |r|} \hline
\textbf{Script} &\textbf{Method} &\textbf{Att.} &\textbf{WES} &\textbf{INIT} &\textbf{WER}$\downarrow$ \\ \hline
Bengali &CNN-RNN &- &- &- &15.23 \\
        &CNN-RNN &$\checkmark$ &  & &14.02 \\ 
        &CNN-RNN &$\checkmark$ &$\checkmark$ & &13.48 \\
        &CNN-RNN &$\checkmark$ &  &$\checkmark$ &13.01 \\
        &CNN-RNN &$\checkmark$ &$\checkmark$ &$\checkmark$ &12.34 \\\hline
\end{tabular}
\end{center}
\caption{Show the performance comparison between different strategies. Attn. represents attention module in the decoder. WES indicates word embedding supervision. INIT represents initializing the state of the GRU in the decoder. $\downarrow$ indicates the lower value corresponds to better performance. \label{table_ablation_study}} 
\end{table}

When attention is used in the decoder, the model reduces the {\sc wer} by 1.21\%. Only the model supervised with word embedding may not reduce much in {\sc wer} (0.54\%). Using predicted holistic features from the encoder to initialize the decoder reduces the {\sc wer} by almost (1.01\%). A combination of word embedding supervision and initialization of the decoder with predicted semantic information reduces most {\sc wer}. 

\paragraph{\textbf{Comparison with State-of-the-art}} We compare our method with state-of-the-art methods on benchmark dataset --- {\sc iiit-indic-hw-words}~\cite{dutta2018offline,dutta2018towards,gongidi2021iiit}. The results are shown in Table~\ref{table_comparison}. Our method reduces overall more than $2\%$ average (average over ten languages) Word Error Rate ({\sc wer}) as compared to the state-of-the-art. Among all languages, due to the complexity of Urdu script, the proposed method has the highest error rate ({\sc wer} and {\sc cer}) for Urdu languages. The proposed method achieves the minimum error rate in the Kannada language.     

\begin{table}[ht!]
\addtolength{\tabcolsep}{7.0pt}
\begin{center}
\begin{tabular}{|l|r r |r r|} \hline
\textbf{Script} &\multicolumn{4}{c|} {\textbf{Performance Score}} \\ \cline{2-5} 
                &\multicolumn{2}{c|} {\textbf{Method~\cite{gongidi2021iiit}}} &\multicolumn{2}{c|}{\textbf{Ours}} \\ \cline{2-5}  
                &\textbf{WER}$\downarrow$ &\textbf{CER}$\downarrow$ &\textbf{WER}$\downarrow$ &\textbf{CER}$\downarrow$ \\ \hline
Bengali         &14.77       &4.85        &12.34     &2.35   \\ 
Gujarati        &11.39       &2.39        &09.21     &1.19   \\
Hindi           &11.23       &3.17        &09.16     &1.98   \\
Kannada         &10.93       &1.79        &08.57     &1.01   \\ 
Malayalam       &11.34       &1.92        &09.37     &1.12    \\ 
Odia            &14.97       &3.00        &12.32     &1.32    \\ 
Punjabi         &12.78       &3.42        &10.77     &2.10    \\ 
Tamil           &11.36       &1.92        &09.18     &1.25    \\ 
Telugu          &13.98       &3.18        &12.11     &2.15    \\ 
Urdu            &20.35       &5.07        &18.76     &3.89   \\ \hline
\end{tabular}
\end{center}
\caption{Quantitative comparison with State-of-the-arts. $\downarrow$ indicates the lower value corresponds to better performance. \label{table_comparison}} 
\end{table}

\begin{figure}
\centerline{
\includegraphics[width=1.0\textwidth]{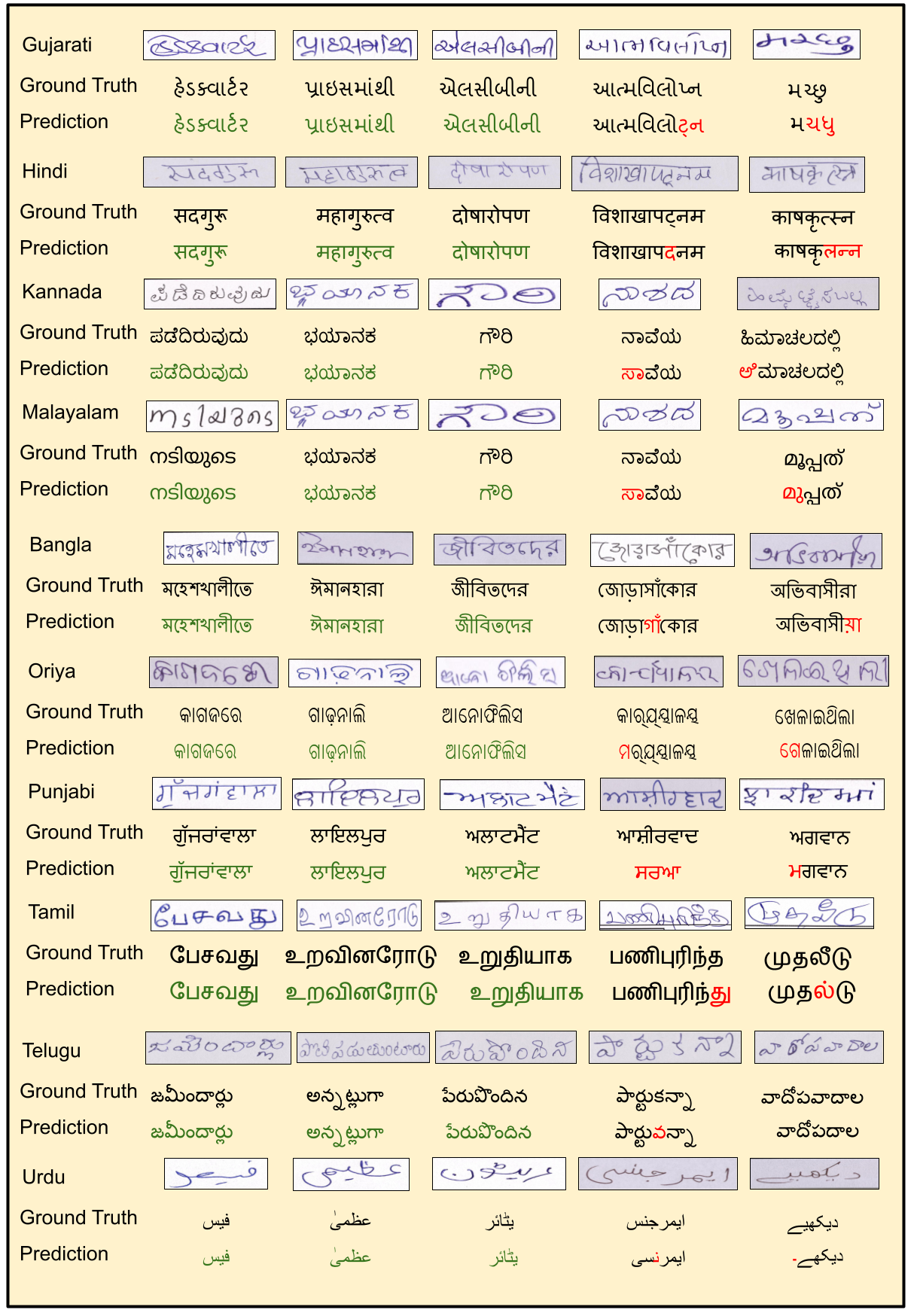}}
\caption{Shows qualitative results obtained by the proposed method. Column 1, 2 and 3 present correctly recognized word images. Column 4 and 5 shows incorrect recognized word images.}\label{fig:visual_result}
\end{figure}

\begin{figure}[!ht]
\centerline{
\includegraphics[width=1.0\textwidth]{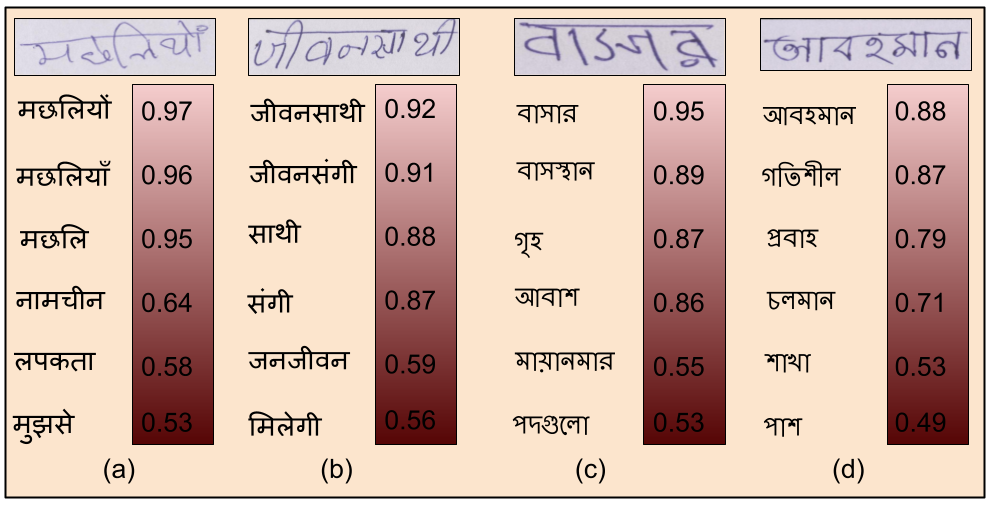}}
\caption{Shows the visualization of cosine similarity between the predicted semantic information from the image and the word embedding of the words from lexicons. Larger value indicates more similar semantics.}\label{fig:visual_embedding}
\end{figure}

\paragraph{\textbf{Qualitative Results and Visualization}}

The visual results shown in Fig.~\ref{fig:visual_result} highlight that the proposed method obtains correct prediction for word images with incomplete characters and blurring characters due to varying ink density. We explain that semantic information will provide an effective global feature to the decoder, robust to the interference in the images. The first three columns of Fig.~\ref{fig:visual_result} show the correctly recognized text of ten languages. While last two columns of Fig.~\ref{fig:visual_result} shows wrongly recognized words by the proposed method. One or more wrongly recognized characters of the words are highlighted by red color. We perform experiments on the {\sc iiit-indic-hw-words} dataset to visualize the validation of the predicted semantic information to the decoder for recognizing text.  
As presented in Fig.~\ref{fig:visual_embedding}, we compute the cosine similarity between the
predicted semantic information from the word image and the word embedding of each word from lexicons (50 words for each image). In Fig.~\ref{fig:visual_embedding}, the predicted semantic information is related to word images which have similar semantics. Words of similar meanings have a large cosine similarity value, while words of different meanings have less cosine similarity value. For example, in the case of Fig.~\ref{fig:visual_embedding} (a), three words from the top have a similar meaning, resulting in a higher cosine similarity score, While the remaining words are different meanings indicating a lower cosine similarity score. A similar observation is found for other word images shown in Fig.~\ref{fig:visual_embedding}. With the help of a semantic module, the model tries to distinguish words very easily. 

\section{Conclusions and Future Works}

This article proposes an Indic handwritten text recognizer using an attention-based encoder-decoder framework with a semantic module to recognize text accurately. The semantic module predicts global semantic information supervised by word embedding from a pre-trained language model. The predicted global semantic information initializes the decoder to recognize the text accurately during decoding for word images having incomplete and blurred characters. The used benchmark dataset contains only word-level images and their ground truth transcriptions. In the future, we will concentrate on Indic handwritten text line recognition by providing datasets and recognizers. It may happen that getting real handwritten documents and manually ground truth transcription generation is time-consuming and cost-ineffective. In the future, we will create synthetic handwritten documents to reduce the time and cost of generating a sufficient amount of real training data.             

\bibliographystyle{splncs04}

\end{document}